\documentclass[letterpaper,twocolumn,10pt]{article}

\usepackage{usenix}
\usepackage{xspace}
\usepackage{xcolor}
\usepackage{pbox}
\usepackage{amsmath}

\usepackage{graphicx}
\graphicspath{ {./img/} }

\usepackage{subfigure}

\usepackage{cleveref}

\usepackage[compact]{titlesec}
\usepackage[font=small, skip=1pt]{caption}

\usepackage{multirow}
\usepackage{array}

\begin{document}

\newcommand{\sys}{Doraemon\xspace}

\date{}

\title{\Large \bf Learned Indexes for Dynamic Workloads}

\author{
    Chuzhe Tang, Zhiyuan Dong, Minjie Wang $^\dag$, Zhaoguo Wang, Haibo Chen \\ 
    {\vspace{0.05in} Shanghai Jiao Tong University, $\dag$ New York University}
}

\maketitle

\begin{abstract}
The recent proposal of learned index structures opens up a new perspective
on how traditional range indexes can be optimized. However, the current learned
indexes assume the data distribution is relatively static and the access
pattern is uniform, while real-world scenarios consist of skew query distribution 
and evolving data. In this paper, we demonstrate that 
the missing consideration of access patterns and dynamic data distribution notably hinders
the applicability of learned indexes. To this end, we propose solutions for learned indexes for
dynamic workloads (called {\sys}). To improve the latency for skew queries, 
\sys augments the training data with access frequencies.
To address the slow model re-training when data distribution shifts,
\sys caches the previously-trained models and incrementally fine-tunes
them for similar access patterns and data distribution.
Our preliminary result shows that, \sys improves the query latency
by 45.1\% and reduces the model re-training time to 1/20.
\end{abstract}

\section{Introduction}


The pioneer study~\cite{kraska2018case} on learned index structures
arouses a lot of excitements around how machine learning can resculpt
system components that have been decades-old, such as bloom filters~\cite{mitzenmacher2018model},
join queries~\cite{krishnan2018learning} or even enable
self-tuning databases~\cite{kraska2019sagedb}.

The core insight of learned indexes is to view index as a distribution function
from the keys to the index positions that can be approximated by
deep neural networks. Nevertheless, their preliminary study assumes
a relatively static distribution function,
while in many real world scenarios, the data is constantly evolving~\cite{council2008tpc}.
Typical approaches simply rely on re-training the whole model once
the data distribution shifts notably from the training set
used by the current model. However, such re-training is costly, because
not only the model parameters need to be \textit{fine-tuned}, but also that the model architecture needs
to be searched again for better accuracy. Depending on the size of
the hyperparamter search space, a basic architecture search technique such as
grid search can easily take up to 10-100x the model training time
\cite{becsey1968nonlinear, lavalle2004relationship, bergstra2011algorithms}.

Besides the inefficiency in handling dynamic workloads, the learned index
paper also assumes a uniform access pattern (or query distribution). However,
queries in real worlds tend to be skew, where some keys are much more frequently
queried than the others \cite{zhang2016reducing, debrabant2013anti, eldawy2014trekking, levandoski2013identifying}. As a result,
mispredicting a hot key is way more expensive, and we show that the originally proposed
learned index model performs poorly under such scenarios. 
These two issues hinder the wider adoption of the learned indexes for real-world workloads.

In this paper, we propose \sys, a new learned index system for dynamic workloads where
the data distribution and access pattern may be skew and evolving. 
To handle skewed access pattern, we first investigate and discuss why the
original model fails to address this issue and then propose an approach that
augments the training data with access frequencies. For the issue of
model re-training, our insight is that the same model architecture can
be reused for similar data distribution and access pattern. Based on this,
{\sys} caches the trained models and simply fine-tunes them when a similar
input distribution is encountered again. The preliminary 
result shows that, by augmenting dataset with the access frequency, the
best model architecture has 45.1\% performance improvement; by caching and
reusing previous training result, the rebuilding time is reduced to 1/20
(from 40 mins to 2 mins).

\section{Learned Indexes}

\begin{table*}
\begin{footnotesize}
\begin{center}
    \scalebox{1.33} 
    {
        \begin{tabular}{|c|cc|cc|cc|cc|}
            \hline
            \multirow{3}{*}{Dataset} & \multicolumn{8}{c|}{Workload}                                                                                                \\ \cline{2-9} 
                                     & \multicolumn{2}{c|}{Skewed 1} & \multicolumn{2}{c|}{Skewed 2} & \multicolumn{2}{c|}{Skewed 3} & \multicolumn{2}{c|}{Uniform} \\ \cline{2-9} 
                                     & Arch          & Time(ns)      & Arch          & Time(ns)      & Arch          & Time(ns)      & Arch          & Time(ns)         \\ \hline
            D1                       & NN16          & 321           & LIN           & 252           & NN16          & 282           & NN16          & 375          \\
            D2                       & NN8           & 319           & NN8           & 316           & NN8           & 301           & LIN           & 344          \\
            D3                       & LIN           & 293           & LIN           & 281           & LIN           & 278           & LIN           & 350          \\
            D4                       & NN8           & 314           & LIN           & 289           & LIN           & 288           & NN8           & 376          \\ \hline
        \end{tabular}
    }
\end{center}
\caption[XIXI] {
    \label{tab:comparison}
    The best model architecture and the corresponding average search time (in ns) with different
    datasets and workloads.
    }
\end{footnotesize}    
\end{table*}

\begin{figure*}
\begin{center}
    \includegraphics[width=1.9\columnwidth]{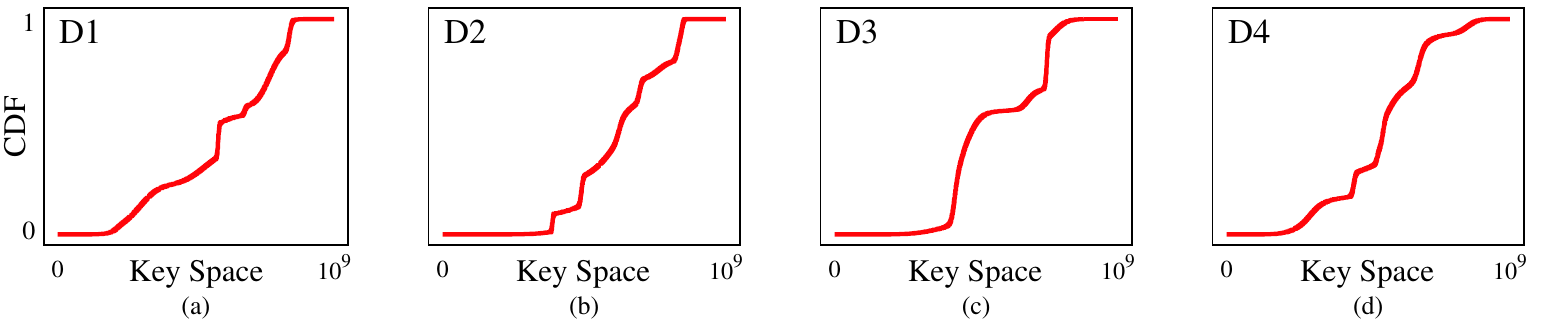}
\end{center}
\caption{
    \label{fig:impact-of-dist-dists}
    Above figures show the CDF of dataset D1, D2, D3 and D4. 
    The y-axis is the normalizd CDF for each dataset.
    The x-axis indicates the key space where key generated from.
}
\end{figure*}


In this section, we introduce the basic background of the original learned
index structures~\cite{kraska2018case}. The insight is that indexes can
be viewed as functions from the data (key) to the values representing either
record positions in a sorted array (for range index), in an unsorted array
(for Hash-Index) or whether the data exists or not (for BitMap-Index).
For the case of range index, the function is effectively a cumulative distribution
function (CDF). Given the CDF $F$, the positions can be predicted by:
\[
p=F(\text{Key})*N
\]
where $p$ is the position of the key and $N$ is the total number of keys 
(see Figure~\ref{fig:impact-of-dist-dists} for examples).

The core idea is to approximate the CDF function $F$ by machine learning models
such as deep neural networks. While the choice of the model architectures
can vary, the paper proposes a \emph{staged model} architecture inspired by
the multi-stage structure of B-Tree. The sub-model at each stage predicts
which sub-models to be activated in the next stage while the leaf stage
directly predicts the CDF values. The model is trained from the root stage
to the leaf stage, and each stage is trained separately using the following loss
function:
\[
L_l=\sum_{(x,y)}(f_l^{(\lfloor M_lf_{l-1}(x)/N\rfloor )}(x)-y)^2~;~L_0=\sum_{(x,y)}(f_0(x)-y)^2
\]
Here, $(x,y)$ is the key/position pair from the data to be indexed;
$L_l$ is the loss function of stage $l$; $f_l^{(k)}$ is the $k^{th}$ sub-model
of stage $l$. $f_{l-1}$ recursively executes the above equation until the root
stage $L_0$.

To deploy the learned index, the approximation error needs to be corrected.
First, the prediction error can be bounded by looking at the maximum distance $\sigma$
between the predicted and the true positions for each key. Hence, if $pos$ is
the predicted position by the learned index, the true position is guaranteed to be
within $[pos-\sigma, pos+\sigma]$, and a binary search can be used. The error
bound $\sigma$ is thus a critical indicator of the effectiveness of the learned
index. The smaller $\sigma$ is, the more effective is the index.

There are several limitations of the original learned index. First, the CDF
should be relatively static. Otherwise, the model needs to be re-trained for
better approximations. Since insertion and deletion are very common, learned indexes
can be quite slow due to the high cost of re-training. Second, the model assumes
all the keys are being uniformly queried, while in reality, the prediction error
of a hotter key has much more impact on the overall performance.

We explain how \sys addresses these issues in the following sections. Section~\ref{sec:challenges}
investigates quantitatively how learned indexes perform under different access patterns (Sec~\ref{sec:pattern})
and data distribution (Sec~\ref{sec:data-dist}). We then propose our solutions in section (Sec~\ref{sec:solution})
using data augmentation (Sec~\ref{sec:augment}) and model caching (Sec~\ref{sec:cache}). We also
discuss other components in our system (Sec~\ref{sec:discuss}) and related works (Sec~\ref{sec:related}).

\section{Challenges with Dynamic Workloads}
\label{sec:challenges}

\begin{figure*}[!hbt]
\begin{minipage}[t]{.5\textwidth}
  \centering
    \includegraphics[scale=0.8]{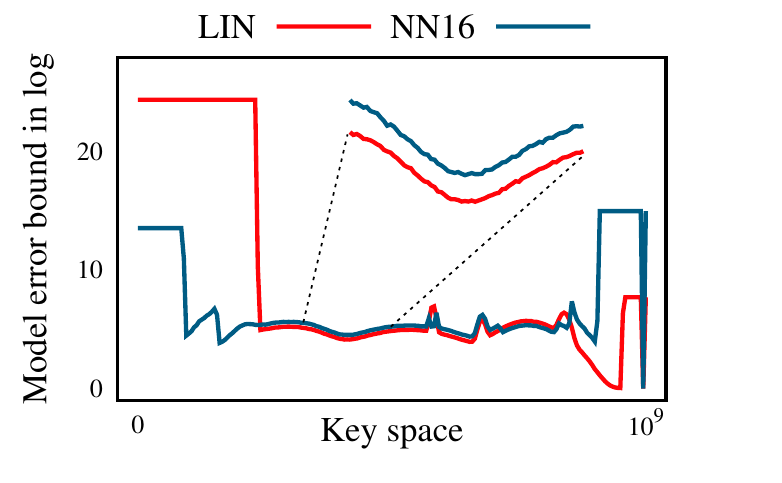}
  \caption{
    \label{fig:errors}
    Errors of 2 configurations. (left) whole key space, (right) hot range of skewed 2. Y-axis is the error bound 
    of the model having the key (X-axis)}
\end{minipage}
~
\begin{minipage}[t]{.5\textwidth}
  \centering
  \includegraphics[scale=0.8]{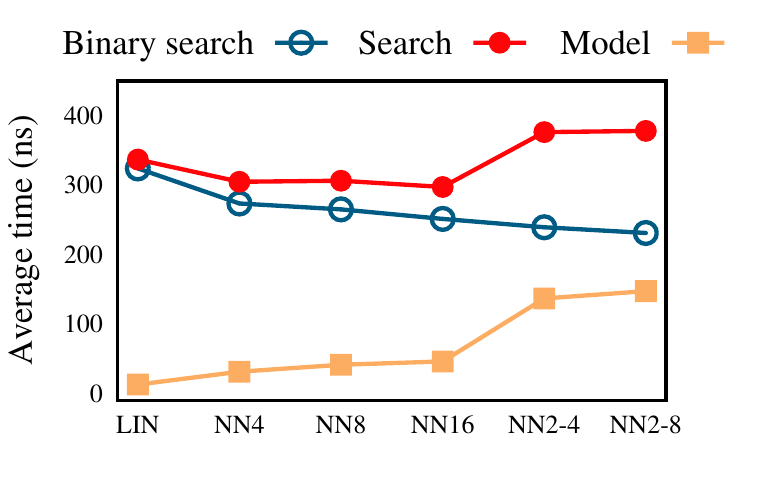}
\caption{
    \label{fig:dist-costs}
    The additional computations of complex models cancel off 
    benefits in binary search time. LIN, NN4/8/16 are defined in Sec 3. NN2-4/8 means two hidden layes NN of width 4 and 8.}
\end{minipage}
\end{figure*}

In this section, we will discuss the challenges posed by dynamic workloads with a simple example of 2 stages learned index. We found that the choice of model architecture is affected by 
both query distribution and data distribution.

\Cref{tab:comparison} compares three different model architectures with
different datasets and workloads. Each dataset has 200M integer keys, 
but with different distributions as shown in~\Cref{fig:impact-of-dist-dists}.
The uniform workload evenly reads every key.
The skewed workloads have 95\% queries reading 5\% hot keys, but in different ranges.
All three architectures have 200k linear models at
the second stage and only their first stages are different.
\begin{itemize}
    \item LIN: The first stage is a linear regression model.
    \item NN8: The first stage is a one hidden layer 8-width Neural Network (NN)
    \item NN16: The first stage is a one hidden layer 16-width Neural Network (NN) 
\end{itemize}

There is an interesting observation based on the results.
By shifting either the workload or the dataset, \emph{the best architecture is 
undecidable}.
For example, for the first row in~\Cref{tab:comparison}, LIN is the best 
with workload Skewed 2, but even worse than B-Tree with workload Skewed 1 (1120 vs. 396 ns).
Next, we will discuss the reasons behind such a phenomenon. 

\subsection{The Query Distribution}
\label{sec:pattern}

Querying a key with learned index has two steps: first, it predicts 
the position by model computation; Second, it tries to find the 
actual position using binary search in a bounded range. However, 
its latency usually depends on the binary search, 
as it takes much longer time than model computation, (6/7--25/26) in 
our evaluation. Further, the search area is decided by the 
error bound\footnote{the difference between minimum and 
maximum prediction error} of the last stage model who has the 
key. Thus, we have the following observation.

\textbf{A skew workload's performance is dominated by 
the hot models' error bound.} Hot model is defined 
as the last stage model who holds a hot (frequently accessed) 
key. Given a workload, all models' error bounds can vary 
across different model architectures, including the hot models'. As a result, 
the best architecture varies for the workloads with different 
query distributions. Figure~\ref{fig:errors} shows the 
error bound (y-axis) of the model where the key (x-axis) 
is located. Two lines represent two architectures, LIN and NN16, 
trained with dataset of D1. For the average 
error bound, NN16's is smaller than LIN's (5.32 vs. 6.58). Thus, 
with uniform workload, NN16 has better performance than LIN 
(375 ns vs. 406 ns). However, for the key range from 3.5$\times$10$^8$ to 
4.6$\times$10$^8$, LIN's average error bound is smaller than NN16's 
(4.56 vs. 4.86). As a result, LIN has better performance than NN16 (252 ns vs. 310 ns) with workload Skewed 2.

\subsection{The Data Distribution}
\label{sec:data-dist}

An advantage of using complex models (e.g., neural networks) at 
the first stage is that it can approximate the complex 
distribution which cannot be fitted with 
linear model. As a result, for those distributions, the complex 
network is able to dispatch the data more evenly than simple 
models, which is good for the uniform workload. For example, 
NN16 is better than both NN8 and LIN (375 ns vs. 390 ns vs. 406 ns) 
for D1 with the uniform workload, as it can appoximate the 
D1 (\Cref{fig:impact-of-dist-dists}.a) more precisely. 

\textbf{Complex model is good for the complex distribution, but 
not always.} This is because of the computation cost 
of complex models. \Cref{fig:dist-costs} shows that with the 
first stage model getting more complex,
even though the binary search time decreases, but the model 
computation time increases. 
Because of this tradeoff, for D3 that exhibits 
relatively complex distribution (\Cref{fig:impact-of-dist-dists}.c), LIN has better performance than NN16 (367 vs. 350 ns) ---
NN16 has better performance than LIN at the binary search 
(317 vs. 336 ns), but it is also penalized by the higher computation cost 
(50 vs. 14 ns).

\section{Proposed Solution}
\label{sec:solution}

\begin{figure}
\begin{center}
    \includegraphics[width=0.95\columnwidth]{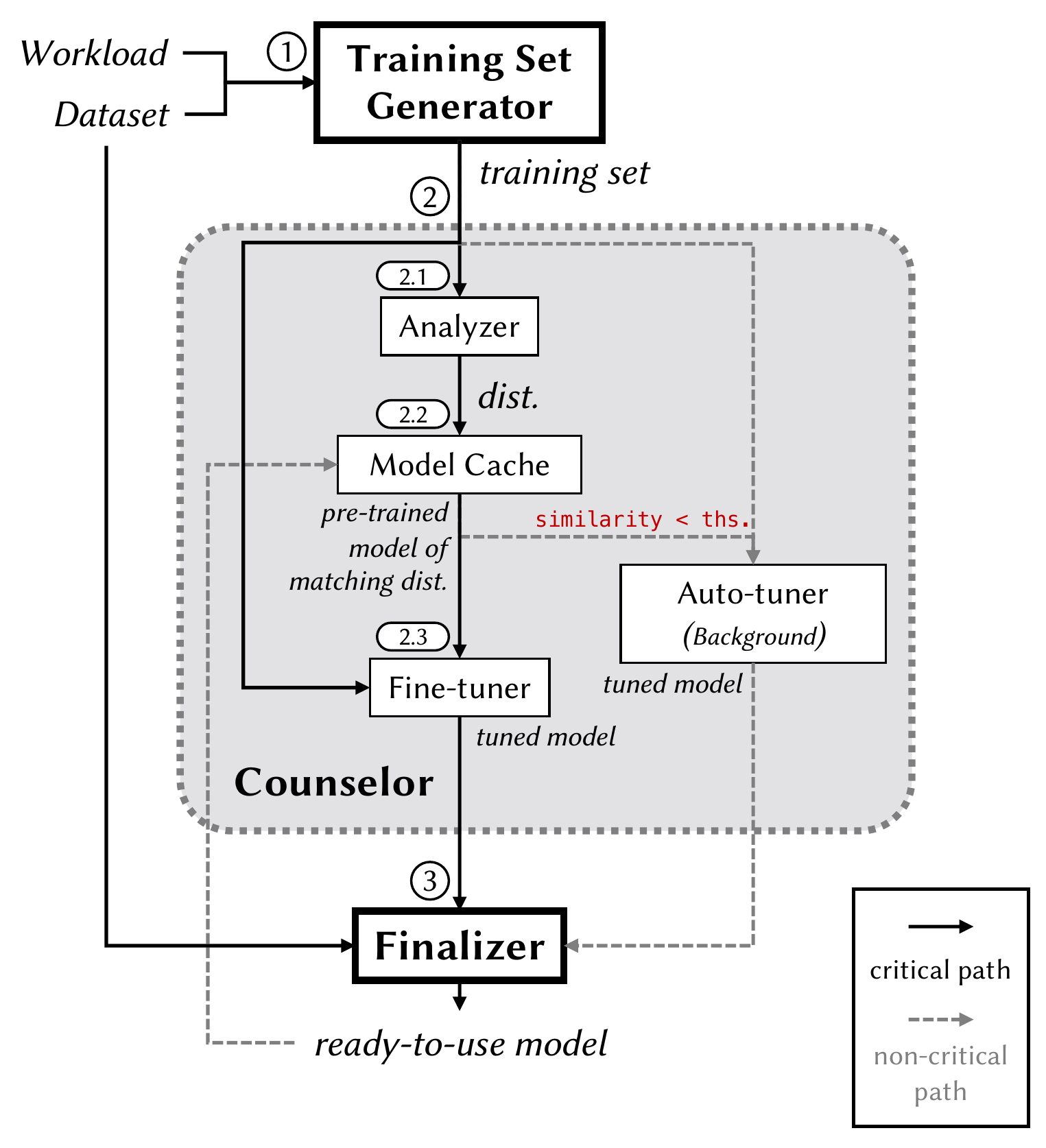}
\end{center}
\caption{
    \label{fig:sys}
    Architecture}
\end{figure}


To achieve learned indexes' best performance, we propose
a new learned index system for dynamic workloads called \sys (\Cref{fig:sys}).
\sys incorporates read access pattern using the \textbf{Training Set Generator} and the
\textbf{Finalizer} and reuses pre-trained models using the \textbf{Counselor}.

\subsection{Incorporate Read Access Pattern}
\label{sec:augment}


To incorporate read access pattern, an intuitive solution is to increase the 
contribution of frequently accessed keys during the training process. 
This can be achieved by creating multiple copies of those keys in the 
training set. For example, considering a training set of 
\{(a, 0), (b, 1), (c, 2)\}, where the first element is 
the key and the second is its position. If the accessed ratio is 1:2:1,
then we double \emph{b} in the training set,
which becomes \{(a, 0), (b, 1), (b, 1), (c, 2)\}. 
In this way, the model will be trained with (b, 1) two times more than others, the prediction 
accuracy of \emph{b} can be improved. We evaluate this intuitive 
solution with the workload of Skewed 3 and the dataset D1. With the new 
training set, the best architecture we can find is NN16 with 275 ns average search time, 
which is close to the previous best architecture, 282 ns. This is because the intuitive solution does not 
improve the error bounds of the second stage models
which decide the search time. In the above evaluation, the 
average error bound does not improve much (5.21 vs. 5.31).

\begin{figure}
    \begin{center}
        \includegraphics[width=0.9\columnwidth]{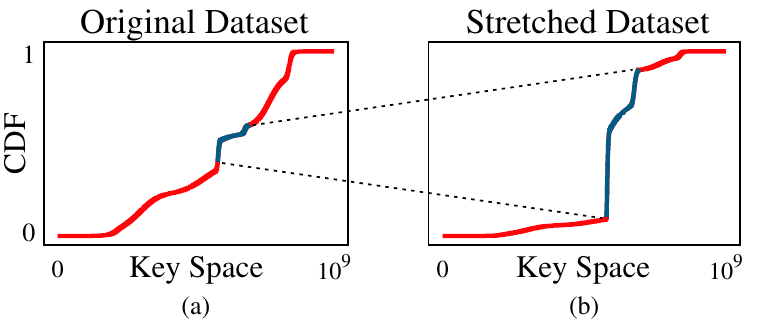}
    \end{center}
    \caption{
        \label{fig:stretch}
        The left figure shows CDF of original D1, while the right figure shows CDF of D1 after stretched.
    }
    \end{figure}
\textbf{``Stretch'' the dataset.} Instead of improving the prediction 
accuracy of the hot keys, we should focus on the error bounds 
of the models containing the hot keys (hot models). Since the models assigned with few keys tend to 
have small error bounds, we try to reduce the number of keys handled by
the hot models by ``stretching'' the dataset.
If a key is frequently accessed, we would like to increase 
the distance between it with its neighbors, the key before or 
after it. It can be achieved by simply shifting the position 
labels. 
Specifically, given a key with position $p$ before ``stretching'', if its access frequency is $f$, and the dataset size is $N$
then we need to shift its position to be $p + (n-1)/2$, and shift 
all keys after it with $n-1$. For the above example, the training 
set of \{(a, 0), (b, 1), (c, 2)\} with access 
frequency 1:2:1 will be augmented to be \{(a, 0), (b, 1.5), (c, 3)\}. Figure~\ref{fig:stretch} shows the CDF of dataset 1 
before and after ``stretching'' with the access pattern in workload 
Skewed 3. 

Training Set Generator takes the workload and dataset as 
input, extracts the access pattern by uniformly sampling 
from the workload and stretches the dataset according to the access 
pattern. Then it sends the stretched training set to Counselor 
to get a tuned model. 

Before using the returned model from Counselor, the Finalizer 
needs to retrain the last stage models with the original dataset. 
This is because the position of each key in the stretched training 
set is changed, we need to repair the position information 
with the original dataset. This process is considerably
fast as last models are usually linear models. For example, 
it only takes 118 $\mu$s to retrain one last model with 1000 keys.

\subsection{Reuse Pre-trained Models}
\label{sec:cache}

After incorporating the access pattern, the only factor affecting
the model architecture is data distribution.
We notice that the best model architecture tends to be the same for similar data distributions.
As a result, \sys is able to cache a mapping from data distributions to 
models for future reusing. 

This is done by the Counselor component, which includes 
four modules: 

\textbf{Analyzer:} extracts distribution information 
by uniformly sampling K records from the generated training set, 
then normalize both key and position to [0, 1]. However, K needs 
to be large enough to avoid 
breaking the distribution. 

\textbf{Model cache:} maintains a mapping 
from the distribution of previous training set to their learning 
model's architure and parameters. If it receives a distribution 
from Analyzer, 
it will finds the entry in the map with the most similar 
distribution 
based on the mean square error. Then, it will send the model's 
information in that entry to Fine Tuner. Furthermore, if the 
similarity 
is below a threshold, it will also start the auto-tuning process. 

\textbf{Fine Tuner:} incrementally trains the model retrieved from 
the model cache with the training set. 

\textbf{Auto-tuner:} uses grid search to find the best model architecture in the given search 
space. 
It performs auto-tuning at the background and sends the result to 
the Finalizer component.

\subsection{Discussion}
\label{sec:discuss}

\textbf{Detecting the change of distribution and access pattern.} 
{\sys} will start to run on detecting the change of distribution or 
access pattern. The detection must be timely with few false positive.
For currently design, we simply detect this by monitoring the 
degradation of the peformance. However, we can use similar technique 
in~\cite{kang2017noscope} to improve the accuracy. 

\textbf{Extract the distribution feature from a dataset.} Currently, we 
simply extract the distribution by uniformly sampling the dataset. 
However, to avoid breaking the distribution, the sample rate varies 
across different dataset. As a result, it is challengin the decide the 
sample rate.

\textbf{Compute the similarity.}
Our sampled distribution representation can be regarded as a type of sequential
data, for which there are many machine learning models are targeting
\cite{greff2017lstm, cho2014learning}.
We believe we can further leverage learning to learn a better similarity metric.

\textbf{Efficiently find the similar distribution in Model Cache.} There can be 
throusands to millions entires in the Model Cache. As a result, 
finding the entry with most similar distribution is considerably 
cost. To solve this issue, we plan to use methods like
\cite{metwally2005efficient, ilyas2008survey} to first filter out the most relevant
entries before the comparison.

\textbf{Improve Auto-tuner efficiency.}
Grid search is slow.
To speed up the search, there are works that use Gaussian process to optimize
the search process \cite{snoek2012practical, bergstra2011algorithms, brochu2010tutorial}.
Similar ideas are also used in database system optimization \cite{duan2009tuning, thummala2010ituned}.

\section{Related Works}
\label{sec:related}

\noindent\textbf{Data Augmentation:} Augmenting training data is a common technique
in machine learning to avoid overfitting and improve generalizability. Many
researchers have been conducted including generating samples through transformation~\cite{baird1992document},
distortion~\cite{simard2003best, yaeger1997effective}, over-sampling~\cite{chawla2002smote}
and from minority class to deal with data
imbalance~\cite{kubat1997addressing, he2008learning, batista2004study}.
As a contrast, the goal of our data augmentation is not to improve generalizability,
but to guide the model to overfit more on keys of high frequency.

\noindent\textbf{Automatic Machine Learning (AutoML):}
Despite the success of machine learning, designing models is still a time-consuming task and require domain expertise.
To ease the problem, many works have been focusing on automatic design and
tuning ML models. Automatic hyperparameter tuning reduces the
tuning efforts by means of grid search~\cite{becsey1968nonlinear,
lavalle2004relationship, bergstra2011algorithms}, random search
\cite{bergstra2012random, bergstra2011algorithms}, Bayesian optimization
\cite{snoek2012practical, brochu2010tutorial}, etc. The search time is usually
proportional to the number of combinations of the hyperparameters to be explored.
Neural Architecture Search (NAS)~\cite{real2018regularized, baker2016designing, zoph2016neural,
liu2018progressive, zhong2017practical, zoph2018learning,
negrinho2017deeparchitect, Zoph_2018_CVPR, Liu_2018_ECCV} are more recent
attempts to design neural network architectures automatically using model-based
optimization strategies such as deep reinforcement learning or progressive search.
These methods usually require tons of computation resources, making it hard
to be deployed in resource-critical scenarios like index lookup directly.

Despite the resource concerns, it is still an open question for AutoML to
handle dynamically-changing data distribution~\cite{automl}. As a result,
in \sys, we combine AutoML with the Model Cache to avoid the costly search for
similar data distribution.

\noindent\textbf{Indexes in Databases}
Indexing is a fundamental component of real-world databases.
Some indexes use hybrid design to serve hot keys and cold keys respectively
by using different data structures, using different storage, and compressing the cold data
\cite{zhang2016reducing, debrabant2013anti, eldawy2014trekking,
levandoski2013identifying}.

Widely used trie-based indexes \cite{leis2013adaptive, mao2012cache} usually
work the best with near uniform data distribution.
To adapt them to less ideal data distributions, Leis et al. use
dynamic fanout to optimize trie height \cite{leis2013adaptive}, Morrison et al. remove unnecessary
nodes \cite{morrison1968patricia}, and Binna et al. aggregate nodes to form a more balanced structure
\cite{binna2018hot}.

\noindent\textbf{Transfer Learning:}
Transferring machine learning models learned from one task to another
different but related task is an active research direction~\cite{pan2010survey,
Dai:2009:EUF:1553374.1553399,pan2011domain}.
Common practice includes reusing learned representations from pre-trained
models and fine-tuning from old weights~\cite{sharif2014cnn,
NIPS2014_5347, donahue2014decaf, oquab2014learning}.
In \sys, models are fine-tuned from the weights obtained from the similar
data distribution, which is easier than transferring models trained from
another distribution.

\noindent\textbf{Data-driven optimizations for system:}
Many system optimizations can be approached by machine learning models trained from historical data.
In the area of database, examples include cardinality estimation
~\cite{lakshmi1998selectivity, kipf2018learned, wu2019towards, park2018quicksel}, join order
planning~\cite{krishnan2018learning, marcus2018deep, ortiz2018learning} and
configuration tuning~\cite{van2017automatic}. Besides database,
works have been done to improve buffer management systems
\cite{chendeepbm}, sorting algorithms \cite{zhao2018n}, memory page prefetching
\cite{hashemi2018learning, zeng2017long} and memory controller \cite{ipek2008self}
and scheduling \cite{kraska2019sagedb}.
Many of these scenarios face similar challenges of dealing with shifting data distribution,
which could be other applications of our model caching mechanism.

\section{Conclusion}
This paper proposes a system which can incorperate the 
query distribution in the training set to improve the query 
performance, and reuse the pre-trained model to reduce the 
re-trained cost. 

\bibliographystyle{plain}
\bibliography{paper}

\end{document}